\documentclass[10pt,twocolumn,letterpaper]{article}

\usepackage{cvpr}
\usepackage{times}
\usepackage{epsfig}
\usepackage{graphicx}
\usepackage{amsmath}
\usepackage{amssymb}
\usepackage{subfig}
\usepackage{booktabs}
\usepackage{multirow}
\usepackage{authblk}

\usepackage[pagebackref=true,breaklinks=true,letterpaper=true,colorlinks,bookmarks=false]{hyperref}

\cvprfinalcopy 

\def\cvprPaperID{****} 

\ifcvprfinal\fi
\begin{document}

\title{Hateful Memes Challenge: An Enhanced Multimodal Framework}
\author[]{Aijing Gao \thanks{agao48@gatech.edu}}
\author[]{Bingjun Wang \thanks{bwang495@gatech.edu}}
\author[]{Jiaqi Yin \thanks{jyin90@gatech.edu}}
\author[]{Yating Tian \thanks{ytian341@gatech.edu \\ Authors ordered by the first name}}
\affil[]{Georgia Institute of Technology}

\maketitle
\begin{abstract}
Hateful Meme Challenge proposed by Facebook AI has attracted contestants around the world. The challenge focuses on detecting hateful speech in multimodal memes. Various state-of-the-art deep learning models have been applied to this problem and the performance on challenge's leaderboard has also been constantly improved. In this paper, we enhance the hateful detection framework, including utilizing Detectron for feature extraction, exploring different setups of VisualBERT and UNITER models with different loss functions, researching the association between the hateful memes and the sensitive text features,  and finally building  ensemble method to boost model performance. The AUROC of our fine-tuned VisualBERT, UNITER, and ensemble method achieves 0.765, 0.790, and 0.803 on the challenge's test set, respectively, which beats the baseline models. Our code is available at: \url{https://github.com/yatingtian/hateful-meme}.
\end{abstract}

\section{Introduction}

\begin{figure}[htp]
    \centering
    \includegraphics[width=0.8\linewidth]{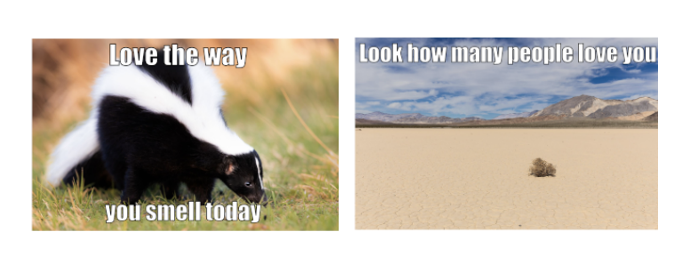}
    \caption{Examples of Hateful Memes \cite{kiela2021hateful}}
    \label{fig:hate-meme-example}
\end{figure}

The Hateful Meme Challenge\cite{kiela2021hateful} is introduced by Facebook AI for multimodal classification, focusing on detecting the hateful speech in multimodal memes. Multimodal problems are trending nowadays yet very challenging. They require a joint understanding of image and text and multimodal reasoning. In the context of multimodal hateful memes on social media platform, it is deemed as hateful when combining text and image. However, while unimodally, both of them are usually considered as harmless. Example memes can be found in Figure \ref{fig:hate-meme-example}. 

Despite multiple multimodal models presented by Facebook such as pretrained VisualBERT\cite{li2019visualbert} and ViLBERT\cite{lu2019vilbert} have shown improvement compared to unimodal models, the gap between the best model and human is still large.  A recent research comparing models of hateful speech detection in multimodal memes and human shows an accuracy of 64.73\% and 84.7\%\cite{velioglu2020detecting}, leaving much room for further improvement.  




There are five data sets from \href{https://hatefulmemeschallenge.com/}{Hateful Memes Challenge}
website:
one train set (\verb|train|), two validation sets (\verb|dev_seen|, 
and \verb|dev_unseen|) and two test sets (\verb|test_seen|, \verb|test_unseen|). For the training set, there are 8500 labeled memes. For validation set, there are 500 labeled memes in \verb|dev_seen| and 540 in \verb|dev_unseen| set. Notice that \verb|dev_unseen| has 400 memes overlapped with \verb|dev_seen|. For test set, there are 1000 \verb|test_seen|
and 2000 \verb|test_unseen| memes, respectively.
In our work, \verb|train|, \verb|dev_unseen|, and
\verb|test_unseen| are used in our training, evaluation, and testing, respectively. For details, please see in Section \ref{sec:method}.


The paper is structured as follows. Section \ref{sec:method} describes the models and the experiment framework. Section \ref{sec:exp_rslt} displays the experiment results. Finally, Section \ref{sec:discussion} summarizes our conclusions and discusses the future work.

\section{Approach} \label{sec:method}

\begin{figure}
    \centering
    \includegraphics[width=0.9\linewidth]{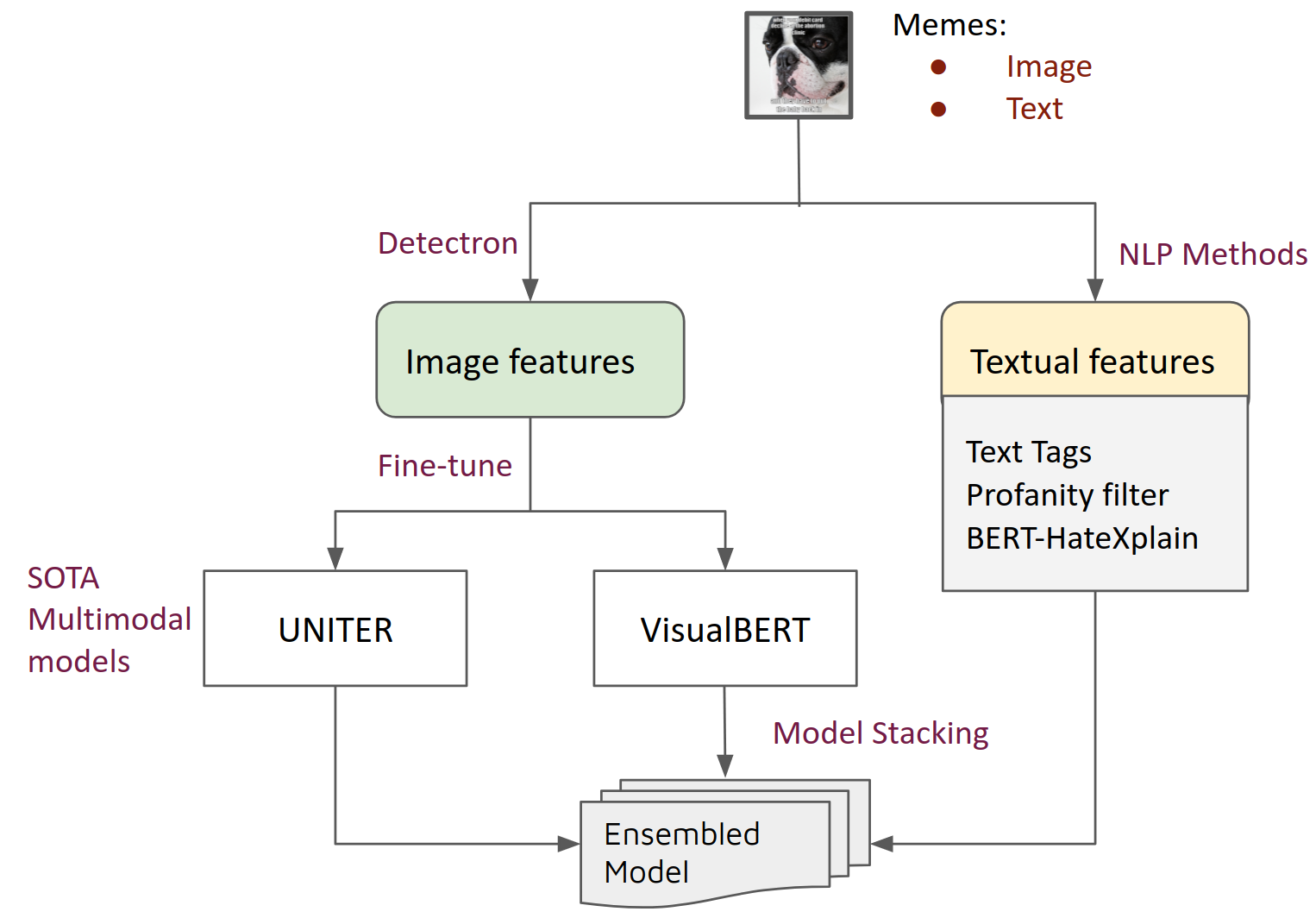}
    \caption{The Complete Deep Learning Pipeline}
    \label{fig:pipeline}
\end{figure}

Our solution to tackle the Hateful Meme Challenge mainly comprises two different transformer architectures: VisualBERT and UNITER. We have applied various methods to construct additional textual and image features and expect to boost the model performance. The complete deep learning pipeline is showed in Figure \ref{fig:pipeline}. The first step is to extract text features and image features. Detectron is used to extract meme image features. For the text feature, a set of sensitive contents showed in meme texts is labeled, such as racism, gender, religion, hateful speech, and etc.. The second step is to use UNITER and  VisualBERT to model the multimodal memes. The final step is to do the hateful meme prediction by combining all the information from the previous steps.

\subsection{Text Feature Extraction}\label{sec:data-processing}

\subsubsection{Tags for Protected Categories}\label{sec:data-protected-category}
Hate speech is usually associated with discrimination based on age, race, color, religion, sex (including sexual orientation, or gender identity), nationality, pregnancy, disability and genetic information. To extend hateful memes with additional labels for identifying those protected categories, we collect a list of keywords from Wikipedia which includes regional nicknames, ethnic slurs, religion slurs, LGBT-related slurs, disability-related terms with negative connotations, and disparaging terms for pregnant women. Through matching meme texts with the list, we create 6 tags that encode details about racism, national origin, sex, religion, pregnancy, and disability.
\subsubsection{Profanity Filter} \label{sec: data-profanity}
We use the \href{https://pypi.org/project/profanity-filter/}{profanity-filter} library to detect profane words in meme texts. The derivative and distorted (e.g. misspelled) profane words are also identified using Levenshtein automata.
\subsubsection{BERT-HateXplain}\label{sec:hate-speech}
We use \href{https://huggingface.co/Hate-speech-CNERG/bert-base-uncased-hatexplain}{BERT-HateXplain}, a BERT-based model pretrained on HateXplain data set \cite{mathew2020hatexplain}, to classify a text as Hatespeech, Normal or Offensive. We observe 65\% of hateful meme texts and 88\% of non-hateful meme texts in training set are classified as Normal.  Similar pattern is found in validation and test set. This shows that text only-based models such as BERT have poor performance compared to multimodal models in tackling this challenge \cite{kiela2021hateful}, possibly due to its limitation in capturing semantic relationship between text and image.  
\subsection{Image Feature Extraction via Detectron} \label{sec:feat-extract}
\begin{figure}
    \centering
    \includegraphics[width=0.8\linewidth]{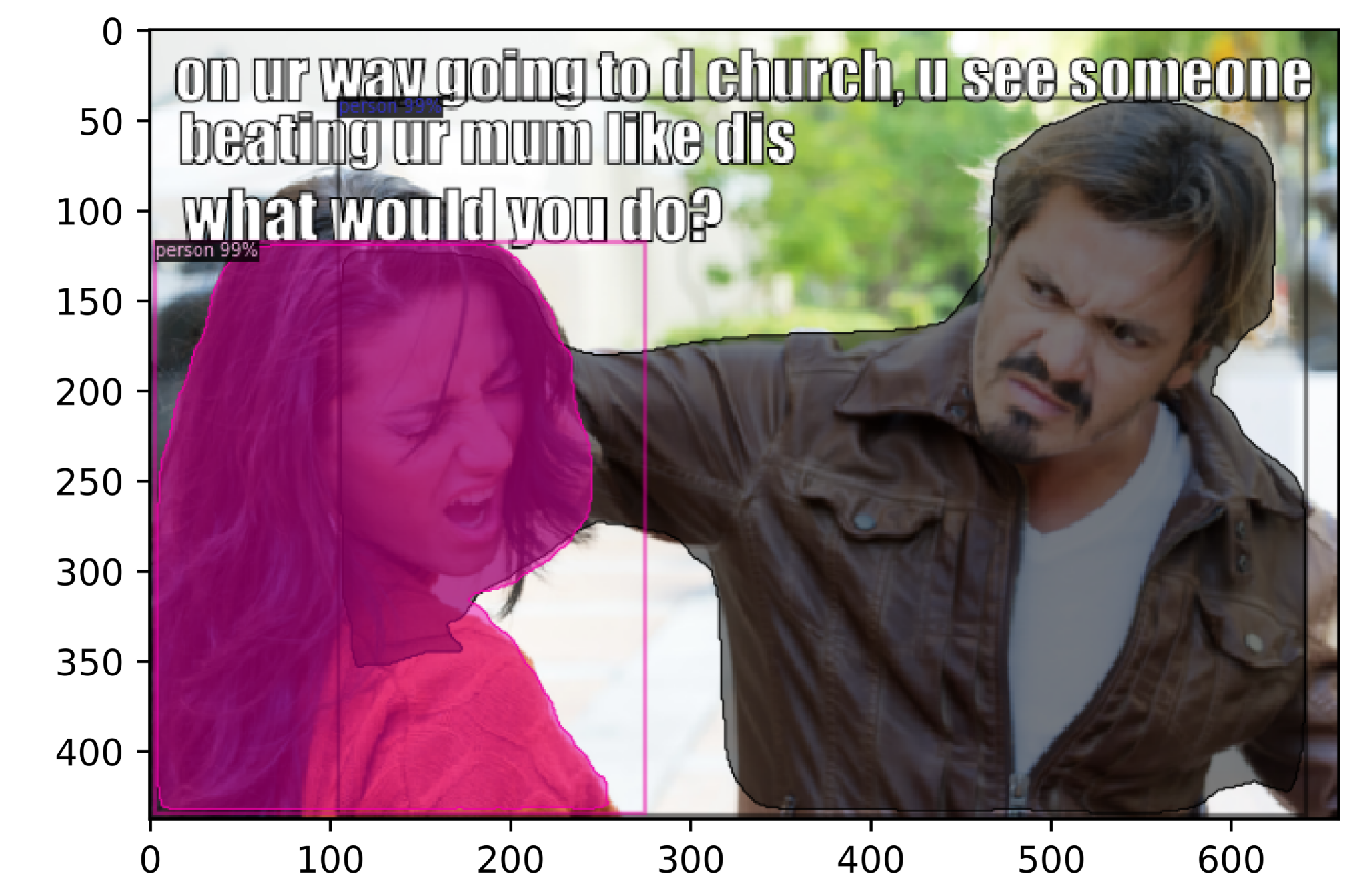}
    \caption{An example of processing hateful meme with Detectron. For visualization purpose, we only show two bounding boxes.}
    \label{fig:detectron}
\end{figure}

To better capture semantic meaning of visual scenes, we perform object detection and construct image features using Facebook's \href{https://github.com/facebookresearch/Detectron}{Detectron}, a Python library that implements multiple state-of-art object detection algorithms. Process of using Detectron for object detection on Hateful Meme could be seen in Figure \ref{fig:detectron}. For each image meme, we extract 120 boxes of 2048D region-based image features using Mask-RCNN with X-152 backbone, which is pretrained on Visual Genome data set.  Those image-based embeddings are projected onto text-based embeddings (768D) before being fed into transformer layers.

\subsection{Losses and Metrics}
Among the training set of 8500 sample size, there are 3019 (35.52\%) labeled as hateful memes, and the rest 5381 (64.48\%) are not hateful. 
Since our training set is imbalanced, besides cross-entropy (CE) loss, we also use focal loss (FL) to give more emphasis on hard and misclassified examples, which are given by the following formula:
\begin{align*}
  \text{CE}(p_t) &= -\log(p_t),\\
    \text{FL}(p_t) &= -(1-p_t)^\gamma \log(p_t).   
\end{align*}    
The performance of our models is determined by area under the receiver operating characteristic curve (AUROC). We also report accuracy (ACC) as a supplemental evaluation metric. Both of them are calculated by using functions in sklearn.metrics.

\subsection{Models}
\subsubsection{VisualBERT} \label{sec:model-visualbert}
VisualBert\cite{li2019visualbert} is a single-stream BERT model with multiple transformer blocks, which projects and converts textual and image embedding into a single embedding before passing into the transformer layers. VisualBERT pretrained on image data sets, including COCO (330K) or Conceptual Captions (CC; 3.3M), has achieved SOTA result in Vision-and-Language tasks such as VQA, Visual Commonsense Reasoning (VCR), and Natural Language for Visual Reasoning (NVLR). Among all baseline models provided by Facebook AI\cite{kiela2021hateful}, VisualBERT with multimodal pretraining on COCO presents the best performance in both validation (AUC: 0.7414) and test set (AUC: 0.7544) of Hateful Memes, indicating its strong ability of finding alignment between text and image. Therefore, we heavily leverage pretrained VisualBERT models in our experiments under Section \ref{sec:exp_rslt}.

\textbf{Pretrained model selection} MMF provides implementation of VisualBERT that is pretrained on a full or reduced-size of COCO or CC data set. Multiple modified implementations are available as well by introducing masked language modeling (MLM) or masked multimodal modeling (MMM)\cite{singh2020pretraining}, where image regions or text inputs are randomly masked with a fixed probability. To identify the best pretrained model on Hateful Memes with Detectron as its feature extractor, we train a list of VisualBERT models with different pretrained keys on a machine containing Tesla P100-PCIE-16GB GPU. We use either cross-entropy or focal loss and evaluate every 50 updates to report the model with the best AUROC score on the validation set. The maximum update is set as 3,000 as we usually find our best model before 3,000 updates. We use the AdamW optimizer with a cosine warmup and cosine decay learning rate scheduler. By default, all those models are initialized from the bert-base-uncased style configuration provided by the HuggingFace library. We also explore the effect of replacing bert-base-uncased style configuration with roberta-large. For all experiments, we use a learning rate of $5e^{-5}$ and batch of size 32. Configuration corresponding to each pretrained VisualBERT model is shown in Table \ref{tab:model-key}. 
\begin{table*}
\centering
\begin{tabular}{lll}
  \hline
 & Pretrained Model & Pretrained Key \\ 
  \hline
 \multirow{2}{4em}{Baseline} & VisualBERT & visual\_bert.finetuned.hateful\_memes.direct \\ 
  & VisualBERT COCO & visual\_bert.finetuned.hateful\_memes.from\_coco \\ 
  \hline
\multirow{3}{4em}{Our Models} & Masked COCO 100\% & visual\_bert.pretrained.coco.full \\ 
 & Masked CC Small 50\% & visual\_bert.pretrained.cc.small\_fifty\_pc \\ 
 & Masked CC Small 10\% & visual\_bert.pretrained.cc.small\_ten\_pc \\ 
   \hline
\end{tabular}
\caption{Finetuned and pretrained models provided by MMF. Baseline models provided by Hateful Meme challenge are fine-tuned with Hateful Memes data. We choose three pretrained visualBERT models with random mask on their input embedding. Masked CC Small 50\% and Masked CC Small 10\% are pretrained on a subset of CC data set.} 
\label{tab:model-key}
\end{table*}

\subsubsection{UNITER}\label{sec:UNITER}


UNITER is a joint multimodal embedding that can be used to bridge the semantics gap between image and text in Vision-and-Language (V+L) tasks. It consists of four pretrained tasks: (1) MLM conditioned on image regions, (2) Masked Region Modeling (MRM) conditioned on input text (with three variants) , (3) Image-Text Matching (ITM), and (4) Word-Region Alignment (WRA).  A multi-layer transformer is then learned from the four pretrained tasks. According to Y.-C. Chen et al.\cite{chen2020uniter}, combining the 4 pretrained tasks (e.g. the UNITER-large model)  could achieve the optimal performance on multiple downstream tasks such as VQA, VCR, NVLR etc.

Since UNITER achieves SOTA performance across various V+L tasks, we include it as part of the solution to solve the hateful meme detection problem. Specifically, the ITM pretrained task of UNITER can be beneficial to our objective as the data set is comprised of 40\% multimodal hate, of which the meaning of the image and text can be counterfactual or contrastive. A model that can unimodally predict well will hardly succeed. Thus, ITM, for image-text alignment, can be reversely used to suit our needs. 

Under the framework of UNITER, we train the model with different number of features. And we consider three different configuration styles of BERT models for initialization. (1) BERT-Large-Cased (Original). This model is trained using the original raw text data without lowercase transformation. 
(2) \href{https://huggingface.co/bert-large-uncased-whole-word-masking}{BERT-Large-Uncased (Whole Word Masking)}\cite{devlin2019bert}. Uncased means the input text is not case-sensitive, for example "english" is treated the same as "English". Different from other pretrained BERT, this model is trained by using a new technique called "Whole Word Masking", which means all the tokens correspond to a word are masked at once. The overall masking rate remains the same. We  found  on  \href{https://github.com/google-research/bert}{Google  Research’s  GitHub  Repo}  that  BERT-Large-Uncased  (Whole  Word  Masking)  is  slightly  better than  BERT-Large-Cased  (Original)  on  The Stanford  Question  Answering (SQuAD  V1.1)  (F1:92.8 vs.  91.5). (3) RoBERTa-Large\cite{liu2019roberta}, which stands for Robustly Optimized BERT Pretraining Approch. It is trained with advanced techniques: dynamic masking, FULL-SENTENCES without NSP loss, large-mini-batches and a large mini-batches. It shows a better performance than BERT-Large on  SQuAD  V1.1  (F1:   94.6  vs.    90.9)  and  some other  data sets.

For the training of UNITER, all tasks are ran on Google Cloud Platform with 1 Nvidia Tesla K80 GPU, 8 vCPUs and 52GB RAM. The models are firstly trained on training set and test on dev\_unseen and then trained on training set and dev\_seen and test on test\_unseen. More details of the model experiments and hyper parameters are discussed in Section \ref{sec: uniter-experiment}.

\subsection{Model Ensemble} \label{sec:method-ensemble}
The idea behind ensemble learning is to combine the predictions of multiple base models to produce a powerful ensemble model, with improved robustness and generalization. We mainly use two types of ensemble methods: 1) Majority vote or average vote that purely combines the predictions of base models. 2) Random Forest that is built upon the predictions of base models while incorporating tags of protected categories, indicator of containing profanity words, and the probability of Hatespeech generated from BERT-HateXplain.


For the first method, we report the AUROC and Accuracy for the \verb|dev_unseen| and \verb|test_unseen| data set.
For the second method, we combine the \verb|dev_unseen| and \verb|dev_seen| (remove the duplicates) as the training data ($N = 640$), with predicted probability of hateful memes from different models and extra text feature information. To find the optimal Random Forest model, the cross validation with random search on the hyper parameters has been implemented. For the Random Forest model, we only report AUROC and Accuracy for \verb|test_unseen| data set. 

More specifically, we compare three different sets of models:
\begin{itemize}
    \item All Model: VisualBERT COCO, Masked COCO 100\% w/ Focal Loss, Masked CC Small 50\%, Masked COCO 100\%, Masked COCO 100\% + RoBERTa, Masked CC Small 10\%, Masked CC Small 50\% w/ Focal Loss, UNITER 36, UNITER 50;
    \item VisualBERT set:  VisualBERT COCO, Masked COCO 100\% w/ Focal Loss, Masked CC Small 50\%, Masked COCO 100\%, Masked COCO 100\% + RoBERTa, Masked CC Small 10\%, Masked CC Small 50\% w/ Focal Loss'
    \item UNITER set:  UNITER 36, UNITER 50.
\end{itemize}

\section{Experiments and Results} \label{sec:exp_rslt}
\subsection{Sensitive Text Tags} \label{sec:sensitive-text-analysis}

\begin{figure}
    \centering
    \includegraphics[width=0.8\linewidth]{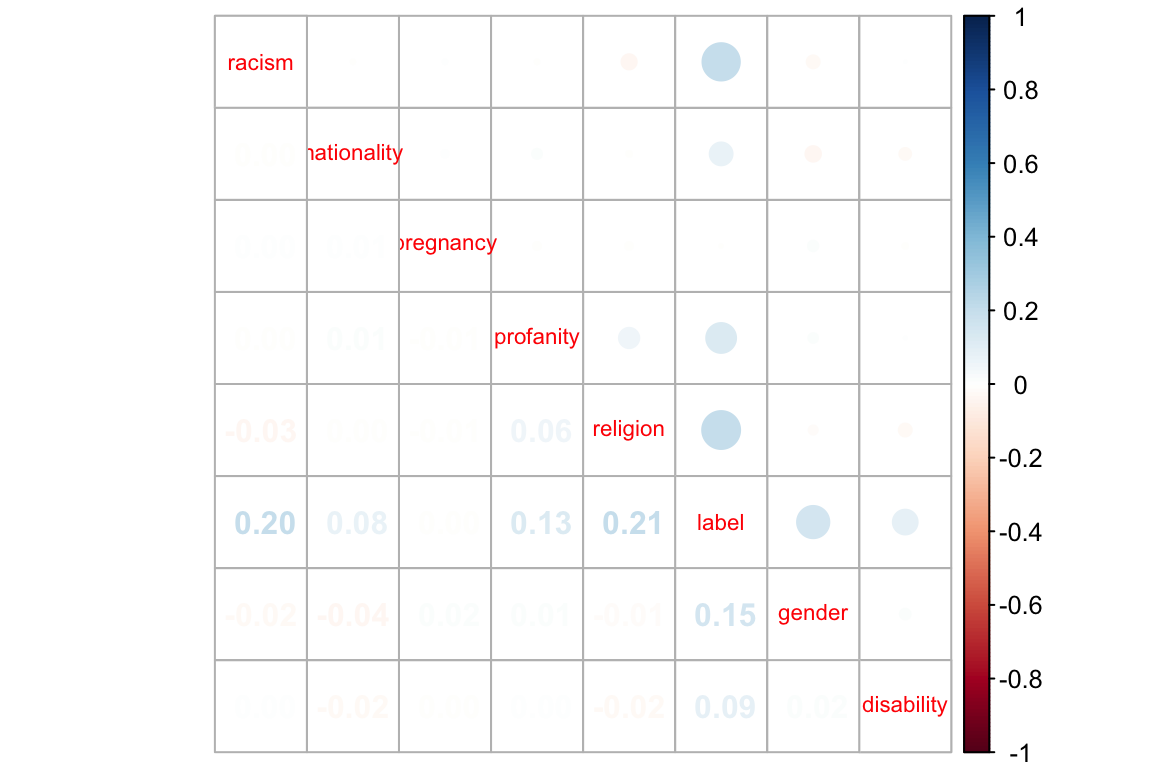}
    \caption{The correlation plot of hateful memes and sensitive tags in the training set. Sensitive contents include words related to racism, nationality, pregnancy, profanity, religion, gender and disability. The lower matrix shows the correlation coefficients.}
    \label{fig:corr-label-sensitive}
\end{figure}

As described in Section \ref{sec:data-protected-category} \& \ref{sec: data-profanity}, we use binary indicator to represent if the memes text has any tags for protected categories or profanity. In the training set, a positive correlation between the hateful memes and the sensitive categories has been shown in  Figure \ref{fig:corr-label-sensitive}. Racism, religion, and gender have relatively higher positive correlation with the hateful memes. In the Table \ref{tab:sensitive-tags}, it displays the number of sensitive categories contained in the meme text with respect to the meme hateful label. There are $793$ memes whose texts have two different categories of sensitive words, and $74.40\%$ ($N = 590$) are labeled as hateful. The percentage of hateful labeling increases to $80.33\%$ and $100\%$ when the number of sensitive categories is 3 and 4, respectively. Such observations encourage us to add the information of sensitive contents in the model ensemble steps; seen in Section \ref{sec:result-ensemble}.

\begin{table}[ht]
\centering
\scalebox{0.9}{
\begin{tabular}{r|ccccc}
  \toprule
& \multicolumn{5}{c}{\# Sensitive Categories}\\ 
\hline 
Hateful Memes & 0 & 1 & 2 & 3 & 4 \\
False & 3284 & 1982 & 203 &  12  &  0 \\
True & 967 & 1411  & 590 &  49  &  2\\
\bottomrule
\end{tabular}
}
\caption{The incidence matrix shows relationship between the meme label and the number of sensitive categories ( racism, nationality, pregnancy, profanity, religion, gender and disability) contained in the meme text. The column is the number of sensitive categories in the each memes and the row the is the meme label.}
\label{tab:sensitive-tags}
\end{table}

\subsection{Finetuning VisualBERT} \label{sec:finetune-visualbert}
We finetune three different pretrained visualBERT models on Hateful Memes with additional 2048D region-based image features  extracted from Detectron. Besides the three pretrained models using default configuration, we also make the following adjustments to the model: 1) Replace cross-entropy loss with focal loss 2) Replace BERT-Base-Uncased style configuration with RoBERTa-large. Each experiment takes about 1.49 to 5.415 hours to complete training. Comparing to baseline performance, adding image region features from Detectron improves AUROC by 1-2\%. We observe that the maximum AUROC score (0.747) on the validation set is Masked COCO 100\% with focal loss. Replacing cross-entropy loss with focal loss in Masked COCO 100\% increases AUROC about 1\% (0.737 vs. 0.747), suggesting mitigating class imbalance could potentially increase model performance. Masked CC Small 50\%, which is pretrained on the half of CC data set, show the best accuracy (72.2\%) and the second largest AUROC score (0.741) among all models. Initializing pretrained weights of RoBERTa-large (AUROC: 0.736) instead of BERT-Base-Uncased (AUROC: 0.737) does not have a significant impact on the model performance of Masked COCO 100\% (in Table \ref{tab:visualbert-rslt}).

\begin{figure}
  \centering
  \subfloat{\includegraphics[width=0.8\linewidth]{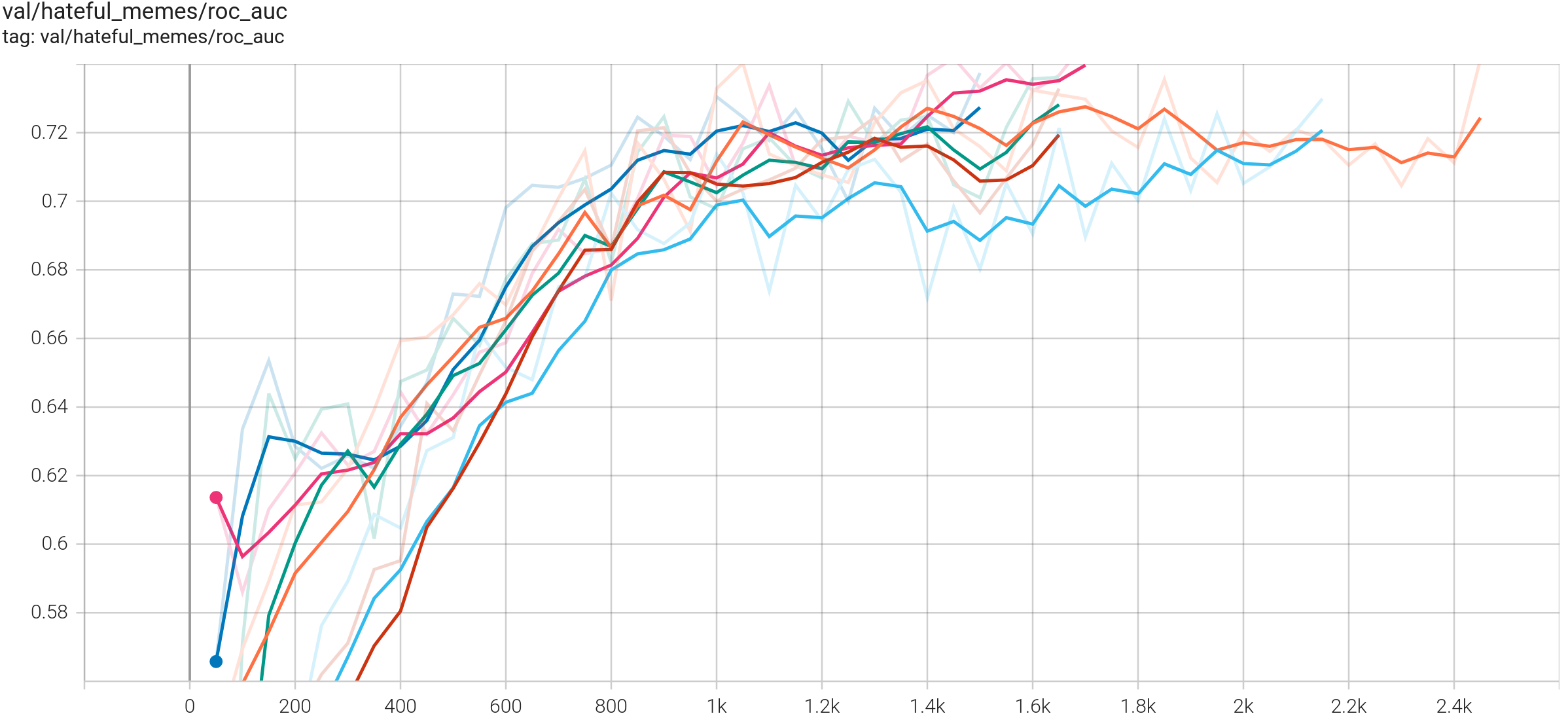}} \\
  \subfloat{\includegraphics[width=0.8\linewidth]{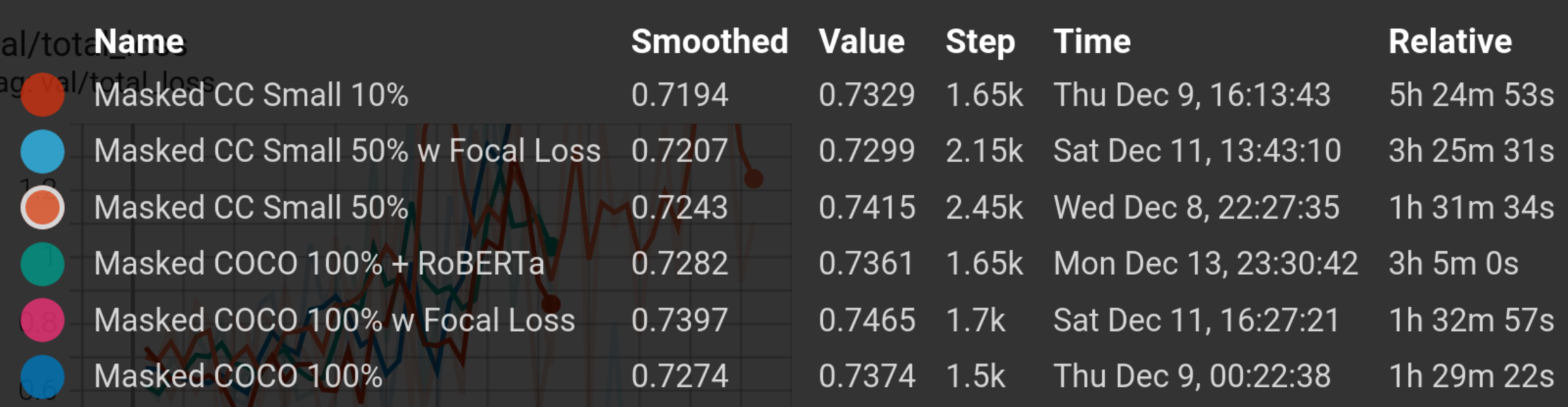}} 
  \caption{Training process of different pretrained VisualBERT models} 
  \label{fig:visualbert}
\end{figure}

Training time to achieve the best performance varies across different models; seen in Figure \ref{fig:visualbert}. With the same experiment set-up, Masked COCO 100\% takes around 1,500 updates to find the best model, significantly faster than other models.   
\begin{table*}[ht]
\centering
\scalebox{0.9}{
\begin{tabular}{l|cc|cc}
  \toprule
 \multirow{2}{*}{Method} & \multicolumn{2}{c|}{Validation} &  \multicolumn{2}{c}{Test} \\
 & AUROC & ACC & AUROC & ACC \\ 
 \hline
Masked COCO 100\% w/ Focal Loss & \textbf{0.747} & 0.719 & 0.755 & 0.713 \\ 
Masked CC Small 50\% & 0.742 & \textbf{0.722} & 0.759 & 0.718\\ 
Masked COCO 100\% & 0.737 & 0.707 & 0.755 & 0.712\\ 
Masked COCO 100\% + RoBERTa & 0.736 & 0.709 &0.756 & 0.720\\
Masked CC Small 10\% & 0.733 & 0.707 & 0.765 & 0.721 \\ 
Masked CC Small 50\% w/ Focal Loss & 0.730 & 0.691 & 0.763 & 0.721\\ 
\hline
VisualBERT COCO (Baseline) & 0.727 & 0.691 & 0.768 & 0.722\\
VisualBERT (Baseline) & 0.679 & 0.661 & 0.731 & 0.695\\
   \bottomrule
\end{tabular}
}
\caption{Using  Detectron  with pretrained  VisualBERT Models on the Validation and Test set} 
\label{tab:visualbert-rslt}
\end{table*}

\subsection{UNITER Vs. Fine-Tuned VisualBERT}\label{sec: uniter-experiment}

This section expands the work of Muennighoff\cite{muennighoff2020vilio}, the second place solution in Hateful Meme Challenge, in which he explored multiple multimodal architectures - OSCAR\cite{li2020oscar}, UNITER, VisualBERT, LXMERT\cite{tan2019lxmert}, and ERNIE-ViL\cite{yu2021ernievil}, and then applied ensemble methods on the output. For UNITER in particular, Muennighoff combined UNITER with BERT-Large-Cased available on \href{https://huggingface.co/bert-large-cased}{Huggingface}, and generated the ensemble results. We experiment the same setup as Muennighoff's and try with different number of features: 36, 50 and 72. Furthermore, we replace the BERT-Large-Cased model configuration in Muennighoff's original UNITER setup with other SOTA Natural Language Processing frameworks, mainly RoBERTa\cite{liu2019roberta} and BERT-Large-Uncased (Whole Word Masking).  

\begin{figure*}[htp]
\centering
\subfloat{
 \includegraphics[width=0.3\linewidth]{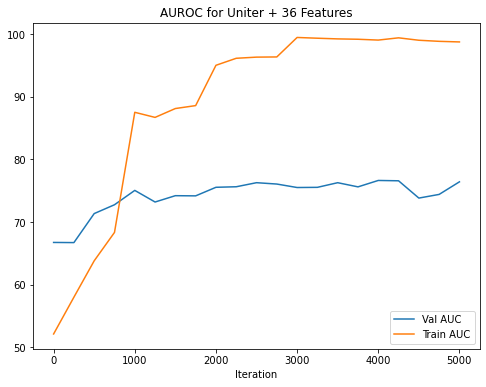}
}
\subfloat{
 \includegraphics[width=0.3\linewidth]{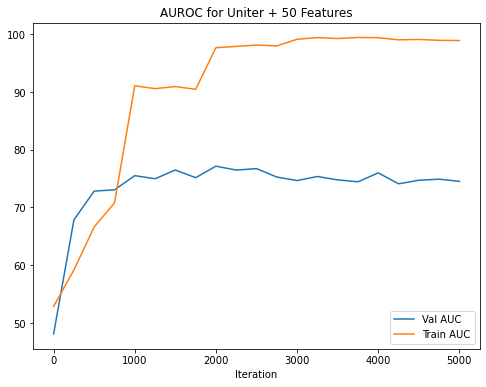}
}
\subfloat{
 \includegraphics[width=0.3\linewidth]{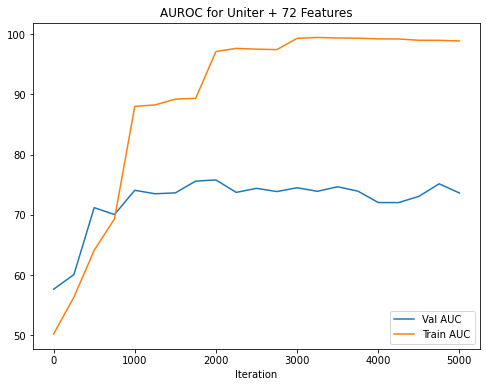}
}
\caption{AUROC for UNITER}
\label{fig:auc-uniter}
\end{figure*}

The performance of UNITER+BERT-Large-Cased with different number of features can be found in Table \ref{tab:uniter-model-output}. The feature extraction method is the same as Section \ref{sec:finetune-visualbert} and described in details under Section \ref{sec:feat-extract}.  The output shows that UNITER with BERT-Large-Cased outperforms UNITER with other experimented NLP frameworks, with 0.780 AUROC on \verb|test_unseen|. Surprisingly, more features do not necessarily lead to a significantly better performance. UNITER 36 achieves slightly better AUROC and Accuracy than UNITER 50 and UNITER 72 on both Validation set and Test set. And due to smaller size of input features, UNITER 36 also has a shorter training time (3.15hrs) than UNITER (3.40hrs) and UNITER 72 (4.29hrs). As seen in Figure \ref{fig:auc-uniter}, AUROC for validation data quickly stabilizes after 1 epoch (1 epoch has 1000 iterations). The figure shows no sign of overfitting for the models.

\begin{table}[ht]
\centering
\scalebox{0.9}{
\begin{tabular}{l|cc|cc}
 \toprule
 \multirow{2}{*}{UNITER} & \multicolumn{2}{c|}{Validation} &  \multicolumn{2}{c}{Test} \\
 & AUROC & ACC & AUROC & ACC \\ 
  \hline
UNITER 36 & \textbf{0.812} & \textbf{0.750} & \textbf{0.790} & 0.740\\ 
UNITER 50 & 0.811 &0.736 & 0.785& \textbf{0.742}\\ 
UNITER 72 & 0.789 & 0.733 & 0.788 & \textbf{0.742} \\ 
UNITER+BERT-Uncased & 0.696 & 0.612 & 0.7022 & 0.684 \\ 
UNITER+RoBERTa  & 0.640 & 0.646 & 0.6151 & 0.641 \\ 
   \bottomrule
\end{tabular}
}
\caption{Finetune UNITER on Hateful Memes data set}
\label{tab:uniter-model-output}
\end{table}

Lastly, we also discover that the better performance of individual model (RoBERTa and BERT-Large-Uncased) does not lead to better performance after combining with UNITER; seen Table \ref{tab:uniter-model-output}. As seen in Table \ref{tab:uniter-model-output}, UNITER+RoBERTa (w/o pretraining) does not achieve comparable result due to the lack of pretraining (with random initialized weights rather than pretrained weights) while all other performing UNITER models utilized pretrained models.  As for BERT-Large-Uncased (Whole Word Masking), we firstly pretrain it with ITM and MLM, with word mask rate of 15\% (for detailed configurations see Appendix \ref{appendix1}), and then load the pretrained model to fine tune UNITER (for detailed configurations see Appendix \ref{appendix1}). In Table \ref{tab:uniter-model-output}, it shows similar results as our baseline models and cannot compete with UNITER+BERT-Large-Cased.

\subsection{Ensemble Learning} \label{sec:result-ensemble}
In Table \ref{tab:ensemble-rslt}, we list the AUROC and Accuracy of validation (\verb|dev_unseen|) and test (\verb|test_unseen|) set  based on different model sets and ensemble methods. The definition of the model set is in Section \ref{sec:method-ensemble}. 
\begin{table}[ht]
\centering
\scalebox{0.9}{
\begin{tabular}{l|cc|cc}
  \toprule
 \multirow{2}{*}{Method} & \multicolumn{2}{c|}{Validation} &  \multicolumn{2}{c}{Test} \\
 & AUROC & ACC & AUROC & ACC \\ 
 \hline
\textbf{All models} & & & & \\
 Major Vote &  0.758 & 0.739 & 0.788 & 0.749 \\
 Average Vote  &\textbf{0.781} & 0.717 & \textbf{0.803}& 0.740 \\
 RF & - & - & 0.802 & 0.668 \\
 \hline
\textbf{VisualBERT}  & & & &\\
   Major Vote  & 0.752& 0.724& 0.783& 0.744\\
 Average Vote & \textbf{0.770}& 0.724 &\textbf{0.790} & 0.746 \\
 RF & - & - & \textbf{0.790}  & 0.610 \\
 \hline
\textbf{UNITER}& & & &\\
   Major Vote & 0.811 & 0.726 & 0.795 & 0.749\\
 Average Vote &\textbf{0.812} & 0.630 & \textbf{0.802} & 0.625 \\
 RF & -& -& 0.762 & 0.655\\
   \bottomrule
\end{tabular}
}
\caption{AUROC and Accuracy after combining different models. RF refers to Random Forecast. THe best AUROC is in bold.} 
\label{tab:ensemble-rslt} 
\end{table}

Overall, the model ensembles have more or less improved the prediction performance, especially the Average Vote. 
When comparing the validation result in Table \ref{tab:ensemble-rslt} with Table \ref{tab:visualbert-rslt}, the AUROC of Major Vote or Average Vote for all three model sets is better than it of any single VisualBERT model. The strength is not obvious on the validation set once adding UNITER models, however, the AUROC is still decent on the test set. One possible reason is that UNITER 36 and UNITER 50 have much better performance compared with VisualBERT models, and once blending them with less competitive models, the AUROC is averaged down. Adding text features into the model does not significantly improve the performance. 


\section{Discussion} \label{sec:discussion}
\subsection{Conclusion}
In this paper, we have explored and compared different SOTA multimodal models with various network architectures, specifically VisualBERT and UNITER. We have finetuned those models by changing number of features, loss function, and configuration style. To boost the performance, we also use different model ensemble methods to aggregate predictions from individual models.

Among all finetuned VisualBERT models, the one pretrained on masked COCO data set with focal loss has the best AUROC (0.747) on the validation set, although their performance on test set is very similar. UNITER significantly outperforms all finetuned VisualBERT models and baseline models for Hateful Meme Challenge, which achieves an AUROC of 0.790  on the test set.

Pretraining is effective in helping to boost the performance of models utilized in our experiments. AUROC of models with pretraining is on-average 8\% better than AUROC of models without pretraining. According to our experiments, more features does not lead to better model performance. With reasonably fewer features, the model can have better performance and training can be done in shorter time frame.

Lastly, model ensemble shows remarkable capability in improving model performance. Once applying model ensemble, the AUROC on the testing set increases from 0.765 in individual VisualBERT model to 0.790.

\subsection{Future Work}
Adding identity tags to both image and text and feeding them as the input features can be a promising direction for future improvement of this task. In our experiments, we extract identity tags (eg. race, gender, sex etc.) from text data and combine them with existing image features in our ensemble step. AUROC has not been improved when adding those text features. The possible reason can be that we only use the text feature tags at the final ensemble step, instead of feeding them into the deep learning training architecture. 

Knowing the context of speech can be important to identify hateful speech, for example text with "Jew" can have a higher probability of hateful speech given the scene presented in the image. The history scene, traditional costume wore by the subject, architectures can convey different metaphors.  Thus, adding knowledge graph embedding to the model can contextualize the background of the image and text. ERNIE-VIL by Baidu Team has the knowledge enhanced vision-language representation through scene graphs, which worth the attention of future studies.

While analyzing the error classification we did in \verb|dev_unseen| set. We found that some of our error is caused by lack in ability to sense information such as race, Hitler and some other specific people on the image of our model. We may need to extract features from images including these critical objects to better resolve this issue.

\section{Other}
The work serves as Fall 2021 Deep Learning (CS-7643) final project at Georgia Institute of Technology.



{\small
\bibliographystyle{ieee_fullname}
\bibliography{egbib}

\begin{thebibliography}{10}\itemsep=-1pt

\bibitem{chen2020uniter}
Yen-Chun Chen, Linjie Li, Licheng Yu, Ahmed~El Kholy, Faisal Ahmed, Zhe Gan, Yu
  Cheng, and Jingjing Liu.
\newblock Uniter: Universal image-text representation learning, 2020.

\bibitem{devlin2019bert}
Jacob Devlin, Ming-Wei Chang, Kenton Lee, and Kristina Toutanova.
\newblock Bert: Pre-training of deep bidirectional transformers for language
  understanding, 2019.

\bibitem{kiela2021hateful}
Douwe Kiela, Hamed Firooz, Aravind Mohan, Vedanuj Goswami, Amanpreet Singh,
  Pratik Ringshia, and Davide Testuggine.
\newblock The hateful memes challenge: Detecting hate speech in multimodal
  memes, 2021.

\bibitem{li2019visualbert}
Liunian~Harold Li, Mark Yatskar, Da Yin, Cho-Jui Hsieh, and Kai-Wei Chang.
\newblock Visualbert: A simple and performant baseline for vision and language,
  2019.

\bibitem{li2020oscar}
Xiujun Li, Xi Yin, Chunyuan Li, Pengchuan Zhang, Xiaowei Hu, Lei Zhang, Lijuan
  Wang, Houdong Hu, Li Dong, Furu Wei, Yejin Choi, and Jianfeng Gao.
\newblock Oscar: Object-semantics aligned pre-training for vision-language
  tasks, 2020.

\bibitem{liu2019roberta}
Yinhan Liu, Myle Ott, Naman Goyal, Jingfei Du, Mandar Joshi, Danqi Chen, Omer
  Levy, Mike Lewis, Luke Zettlemoyer, and Veselin Stoyanov.
\newblock Roberta: A robustly optimized bert pretraining approach, 2019.

\bibitem{lu2019vilbert}
Jiasen Lu, Dhruv Batra, Devi Parikh, and Stefan Lee.
\newblock Vilbert: Pretraining task-agnostic visiolinguistic representations
  for vision-and-language tasks, 2019.

\bibitem{mathew2020hatexplain}
Binny Mathew, Punyajoy Saha, Seid~Muhie Yimam, Chris Biemann, Pawan Goyal, and
  Animesh Mukherjee.
\newblock Hatexplain: A benchmark dataset for explainable hate speech
  detection, 2020.

\bibitem{muennighoff2020vilio}
Niklas Muennighoff.
\newblock Vilio: State-of-the-art visio-linguistic models applied to hateful
  memes, 2020.

\bibitem{singh2020pretraining}
Amanpreet Singh, Vedanuj Goswami, and Devi Parikh.
\newblock Are we pretraining it right? digging deeper into visio-linguistic
  pretraining, 2020.

\bibitem{tan2019lxmert}
Hao Tan and Mohit Bansal.
\newblock Lxmert: Learning cross-modality encoder representations from
  transformers, 2019.

\bibitem{velioglu2020detecting}
Riza Velioglu and Jewgeni Rose.
\newblock Detecting hate speech in memes using multimodal deep learning
  approaches: Prize-winning solution to hateful memes challenge, 2020.

\bibitem{yu2021ernievil}
Fei Yu, Jiji Tang, Weichong Yin, Yu Sun, Hao Tian, Hua Wu, and Haifeng Wang.
\newblock Ernie-vil: Knowledge enhanced vision-language representations through
  scene graph, 2021.

\end{thebibliography}
}

\newpage

\section{Appendix}

\subsection{Hyperparameters used in Section 3.4}\label{appendix1}

\textbf{UNITER+BERT-Large-Cased w/ Different No. Feats}

\begin{itemize}
  \item Epcoh = 5
  \item Batchsize = 8
  \item lr = 1e-5
  \item optimizer = AdamW
  \item Dropout = 0.1
\end{itemize}

\textbf{UNITER+BERT \& UNITER+RoBERTa}
\begin{itemize}
  \item Epcoh = 5
  \item Batchsize = 8
  \item lr = 1e-5
  \item optimizer = AdamW
  \item Dropout = 0.1
\end{itemize}

\textbf{UNITER+BERT-Large-Uncased}
\begin{enumerate}
  \item pretraining
    \begin{itemize}
      \item Epcoh = 8
      \item Batchsize = 8
      \item lr = 0.25e-5
      \item optimizer = AdamW
      \item Dropout = 0.1
      \item WordMaskRate = 0.15
    \end{itemize}
  \item Training
    \begin{itemize}
      \item Epcoh = 5
      \item Batchsize = 8
      \item lr = 1e-5
      \item optimizer = AdamW
      \item Dropout = 0.1
    \end{itemize}
\end{enumerate}

\end{document}